\documentclass[11pt]{article}
\usepackage[margin=1in]{geometry}
\usepackage[T1]{fontenc}
\usepackage{lmodern}
\usepackage{microtype}
\usepackage{amsmath,amssymb}
\usepackage{booktabs,graphicx}
\usepackage[hidelinks]{hyperref}
\usepackage{enumitem}
\setlist{nosep}
\title{When Facts Change, Which Decisions Break?\\Dependency-Aware Context Graphs for Agents}
\author{Paul Deepakraj Retinraj\\Scale AI\\\texttt{paul.retinraj@scale.com}}
\date{}
\begin{document}
\maketitle
\begin{abstract}
When an observed fact changes, an agent must identify and repair dependent decisions while preserving unaffected outputs. We study the reliability--cost trade-off of field-level dependency graphs extracted from actual agent read traces. A controlled benchmark spans three executable workflow domains and four successive updates, with separate imperative and expression-based references. Across 120 workflows, two gateway models, and two repeated runs, we compare full recomputation, rolling summary memory, lexical retrieval, trace-extracted dependencies, and verified dependency upper bounds. Trace-derived graphs reduce review work but can omit unvisited alternatives and necessary repair context. Paired ablations examine edge deletion, spurious edges, transitive propagation, and context scope. An exploratory enhancement adds the observable record ID/field directory during extraction without revealing unread values or reference edges. On 60 fresh workflows with two repeats, the enhanced graph matches observed full-context complete-plan correctness for gpt-6.1-sol (100\%) with an estimated 4.0\% total-cost reduction, but reaches only 84.8\% versus 99.6\% for gpt-6-luna. The findings motivate evaluating selective repair with executable correctness and full memory-processing cost together. The evidence concerns explicit procedural workflows and does not establish broad real-world agent-memory generalization.
\end{abstract}

\section{Introduction}
Agent decisions can remain in memory after the facts supporting them change. A supplier selected under an earlier delivery estimate may become ineligible; downstream spend, approval, and reminder decisions can then become stale. Revising a fact alone is insufficient, while recomputing every output may incur unnecessary work. Explicit dependencies offer a way to select candidate decisions for review, but their usefulness depends on the completeness of the extracted support structure and the evidence available during repair.

We investigate this trade-off with a prototype context graph and a controlled executable workflow benchmark. The central question is not whether reverse reachability identifies graph descendants, but whether a graph extracted from an agent's actual observations supplies enough information to maintain a correct plan across successive updates. An initially sufficient read trace need not describe every alternative that can become relevant later. This creates a mismatch between initial planning and future repair.

Our contribution is an empirical evaluation rather than a claim that dependency tracking or cascading invalidation is new. We provide: (i) sequential workflows with executable reference decisions, (ii) comparisons that include initial planning, memory construction, maintenance, and repair cost, (iii) paired dependency and context ablations, and (iv) fresh-workflow evaluation of a simple observable-directory enhancement. We report failures and limitations alongside successes. The results do not support an unconditional replacement of full-context recomputation with extracted graph repair.

\section{Related work}
StateGraph explicitly models state dependencies, verifies counterfactual support, handles alternative support, and evaluates selective cascade revision and public-memory tasks \cite{stategraph}. PACMI studies provenance-aware cascading memory invalidation and repair \cite{pacmi}. These are directly overlapping mechanisms. We do not reproduce either system and cannot claim comparative superiority.

\emph{Impact Is Not Invalidation} distinguishes a changed artifact's impact from whether a stored claim actually becomes false, using executable claim ground truth \cite{impact}. Our candidate sets are likewise separated from decision correctness: reachability is a conservative review signal rather than proof of invalidity.

LongMemEval evaluates long-term interactive memory, including knowledge updates and temporal reasoning \cite{longmemeval}; AMA-Bench studies long-horizon memory for agentic applications and causality-graph-based memory \cite{ama}. Our external LongMemEval check is evidence-only calibration, not full-history evaluation. The narrow framing here is the reliability and total cost of trace-derived field dependencies for executable workflow repair over repeated updates.

\section{Problem and methods}
\subsection{Decisions, changes, and review candidates}
Let $F_t$ denote observable fact records after update $t$, and let $D$ be a fixed set of requested decisions. An explicit policy specifies an executable reference $y_t^*=f(F_t)$, while method $m$ maintains output vector $\hat y_t^{(m)}$. A decision changes in the reference when
\[
 I_t^*=\{d\in D:y_{t,d}^*\ne y_{t-1,d}^*\}.
\]
The stale-output set for a particular method is instead
\[
 I_t^{(m)}=\{d\in D:\hat y_{t-1,d}^{(m)}\ne y_{t,d}^*\}.
\]
These sets can differ because earlier model mistakes remain in the maintained plan. We retain those mistakes rather than replacing them with oracle values.

An edge $d\rightarrow s$ means decision $d$ depends on support $s$, annotated with watched fields. Supports can be fact records or other decisions. A field-level update $\Delta_t$ first selects decisions whose support fields intersect the changed fields, then propagates conservatively through dependent decisions. The resulting candidate set $C_t$ is not necessarily $I_t^*$ or $I_t^{(m)}$: a changed premise can leave an output unchanged, for example when an alternative support still suffices.

The prototype stores nodes and \texttt{DEPENDS\_ON} edges in an in-memory SQLite \texttt{GraphStore}. Its existing \texttt{review\_plan} function performs reverse traversal with field filtering. Graph edges and initial stored node values are built once; current facts enter the repair context externally. This experiment therefore does not evaluate temporal edge maintenance or bitemporal correction.

\subsection{Initial agent and extraction}
The initial agent receives policies and a record directory containing IDs and field names, but not values. It makes an actual \texttt{read\_records} function call, receives the requested observable records, and returns a plan with brief public explanations. Each repair method shares that initial plan within a workflow/model/repetition trial.

The main extractor receives the policies, plan, and actual tool-read results. It is instructed to enumerate direct field dependencies, including alternatives that may become relevant after changes. It does not receive reference expressions or reference dependencies. Unknown IDs, invalid fields, self-supports, and malformed annotations are rejected by the harness. Extraction quality is measured against conservative static expression references, including alternative branches; these labels are not necessary-active-cause annotations.

\subsection{Repair methods}
\begin{description}[style=nextline,leftmargin=0pt]
\item[Full context.] Recompute all decisions using all current fact records and the prior plan.
\item[Rolling summary.] Create a working memory prompted to at most 1,200 characters, update it with each fact patch, and recompute all outputs from that memory. This is a prompted limit, not hard truncation.
\item[Lexical retrieval.] Use a fixed keyword-overlap rule to choose eight decisions and eight fact records. This is a lightweight lexical baseline, not an embedding retriever or a reproduction of a named memory system.
\item[Extracted graph.] Select graph candidates, retrieve their recursively required support fields and stored decision values, and ask the model to recompute the requested outputs together. Unselected outputs are retained.
\item[Verified graph.] Use reference dependency annotations with the same graph selection and repair machinery. This is a diagnostic upper bound with free oracle annotations, not a deployable method with measured annotation acquisition cost.
\end{description}
Missing information can produce null/unknown outputs, which count as incorrect. If the candidate set is empty, no repair call is made. There are no paid SDK retries, deterministic replacement answers, or hidden fallback methods.

\section{Experimental design}
\subsection{Procedural workflows}
The main set contains 120 workflows, 40 per domain: procurement, scheduling, and resource allocation. Procurement includes supplier eligibility, alternative suppliers, spend, approval, ownership, and dependent reminders. Scheduling includes job ordering, sequential start/finish calculations, deadlines, and downstream coordination. Resource allocation includes ordered demand allocation, remaining capacity, coverage, costs, and ownership.

Each workflow includes unrelated fact records and four independent noise decisions. Development and test families differ in numbers of alternatives, reminder chain length, or noise volume. There are only two test families per domain, so parameter instances should not be treated as broad structural diversity. These families are held out from harness development, not model pretraining.

Four successive patches introduce a relevant shock, changed eligibility or order, an irrelevant metadata update, and a reversal or late adjustment. Policies define one exact expected answer. The generator cross-checks separate imperative and expression-DAG evaluators at every snapshot. Both evaluators still share the generator and parameters; there is no independent human annotation audit.

\subsection{Models, repetitions, and cost}
Two OpenAI-compatible gateway identifiers were used: \texttt{openai/gpt-6-luna} and \texttt{openai/gpt-6.1-sol}, abbreviated Luna and Sol below. These identifiers and returned served identities are retained, but dated weight snapshots and public availability are not independently established. Responses use low reasoning effort, a 4,096-token output maximum for main study calls, no server-side storage, and disabled retries.

Each workflow/model pair has two independent runs. The main study therefore contains 480 initial agent trajectories and $480\times5\times4=9{,}600$ method-update evaluations. The initial tool-use and plan-generation calls are shared within a trial but included in each method's hypothetical standalone cost. Costs use runtime gateway prices: Luna input/output \$0.10/\$0.50 per million tokens and Sol \$2.00/\$10.00 per million tokens. Cached-input discounts are ignored. These are rate-based estimates, not provider invoices.

For a deployed method, total trajectory cost is
\[
 K_m=K_{\mathrm{initial}}+K_{\mathrm{memory},m}
       +\sum_{t=1}^{4}(K_{\mathrm{maintenance},m,t}+K_{\mathrm{repair},m,t}).
\]
Graph extraction and summary generation are included rather than amortized away. The four-update horizon does not establish a universal long-run break-even point.

\subsection{Endpoints and uncertainty}
The primary endpoint is complete-plan correctness after each update:
\[
 A_m=\frac1N\sum_{i=1}^{N}\frac1{RT}
       \sum_{r=1}^{R}\sum_{t=1}^{T}
       \mathbf{1}[\hat y_{i,r,t}^{(m)}=y_{i,t}^*],
 \quad R=2,\ T=4.
\]
A single stale, missing, or unknown output makes a plan incorrect. We also report decision accuracy, initial correctness, edge precision/recall, selected outputs, changed-set coverage, tokens, and total estimated cost.

For paired accuracy differences, steps and repetitions are averaged inside a workflow, then 5,000 workflow-cluster bootstrap samples are drawn within domains. Intervals are descriptive 95\% intervals; repetitions and steps are not independent samples. We do not make a multiple-comparison-adjusted familywise significance claim. When every sampled workflow succeeds, bootstrap intervals can be degenerate and do not imply zero population error.

\section{Results}
\subsection{Main comparison}
Table~\ref{tab:main} reports the main comparison. Extracted graphs review substantially fewer outputs, but this reduction is accompanied by lower complete-plan correctness. The verified-graph diagnostic separates annotation limitations from the repair model's ability to use a correct support structure.
\begin{table}[htbp]
\centering\small
\begin{tabular}{llrrrr}
\toprule
Model & Method & Accuracy (\%) & 95\% CI & Reviewed & Total cost (\$)\\
\midrule
gpt-6-luna & Full context & 99.4 & [98.8, 99.9] & 14.67 & 0.001601 \\
gpt-6-luna & Rolling summary & 61.1 & [55.8, 66.4] & 14.67 & 0.002410 \\
gpt-6-luna & Lexical retrieval & 24.8 & [22.0, 27.7] & 8.00 & 0.001167 \\
gpt-6-luna & Extracted graph & 87.6 & [84.1, 90.9] & 5.82 & 0.001553 \\
gpt-6-luna & Verified graph & 98.0 & [96.5, 99.4] & 6.08 & 0.000923 \\
gpt-6.1-sol & Full context & 100.0 & [100.0, 100.0] & 14.67 & 0.027286 \\
gpt-6.1-sol & Rolling summary & 99.6 & [98.8, 100.0] & 14.67 & 0.047122 \\
gpt-6.1-sol & Lexical retrieval & 24.3 & [21.5, 27.2] & 8.00 & 0.021335 \\
gpt-6.1-sol & Extracted graph & 73.3 & [72.0, 74.6] & 5.43 & 0.024503 \\
gpt-6.1-sol & Verified graph & 100.0 & [100.0, 100.0] & 6.08 & 0.015416 \\
\bottomrule
\end{tabular}
\caption{Main study: 120 workflow clusters, two repeats per model, four updates. Reviewed is mean selected outputs per update. Cost includes initial planning, memory construction and maintenance, and repair. Verified annotations are free oracle inputs.}
\label{tab:main}
\end{table}

For gpt-6-luna, extracted graph accuracy differs from full context by -11.8 percentage points (paired 95\% CI [-15.3, -8.4]), with estimated total cost 3.0\% lower. This is not a quality-equivalent saving.

For gpt-6.1-sol, extracted graph accuracy differs from full context by -26.7 percentage points (paired 95\% CI [-28.0, -25.4]), with estimated total cost 10.2\% lower. This is not a quality-equivalent saving.

\begin{figure}[htbp]
\centering\includegraphics[width=\linewidth]{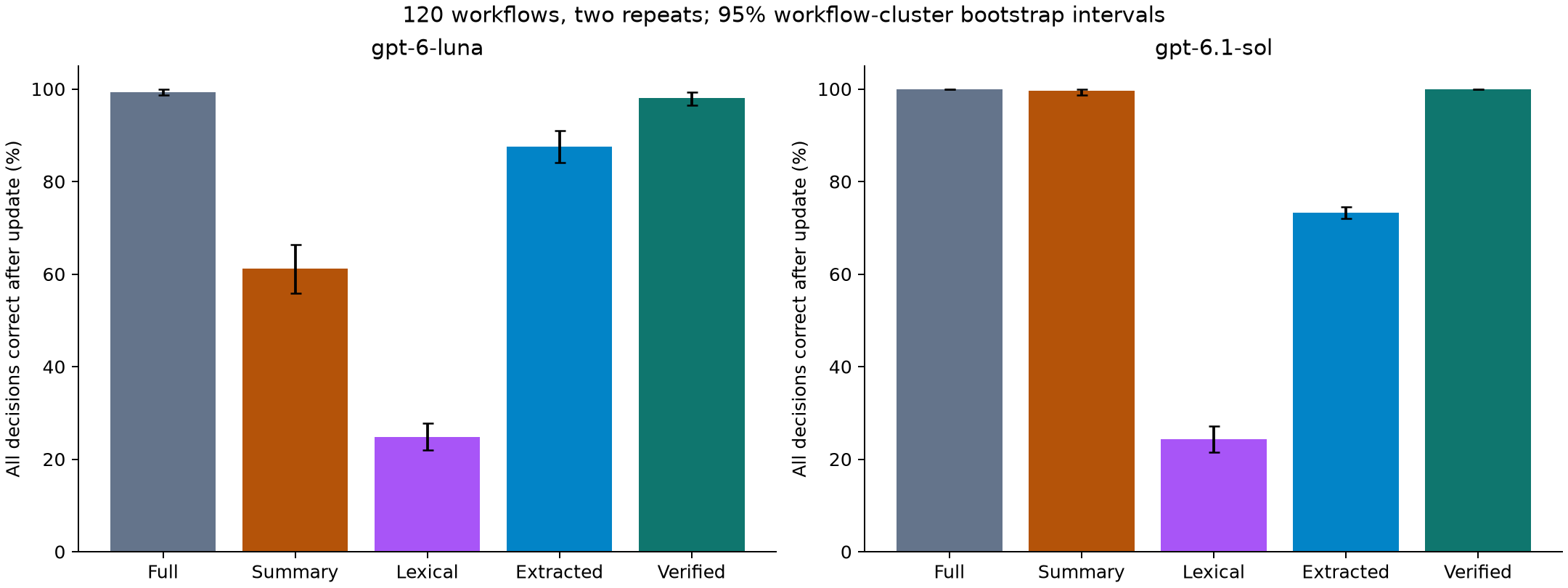}
\caption{Complete-plan correctness in the main study. Error bars are workflow-cluster bootstrap intervals.}
\end{figure}
\begin{figure}[htbp]
\centering\includegraphics[width=\linewidth]{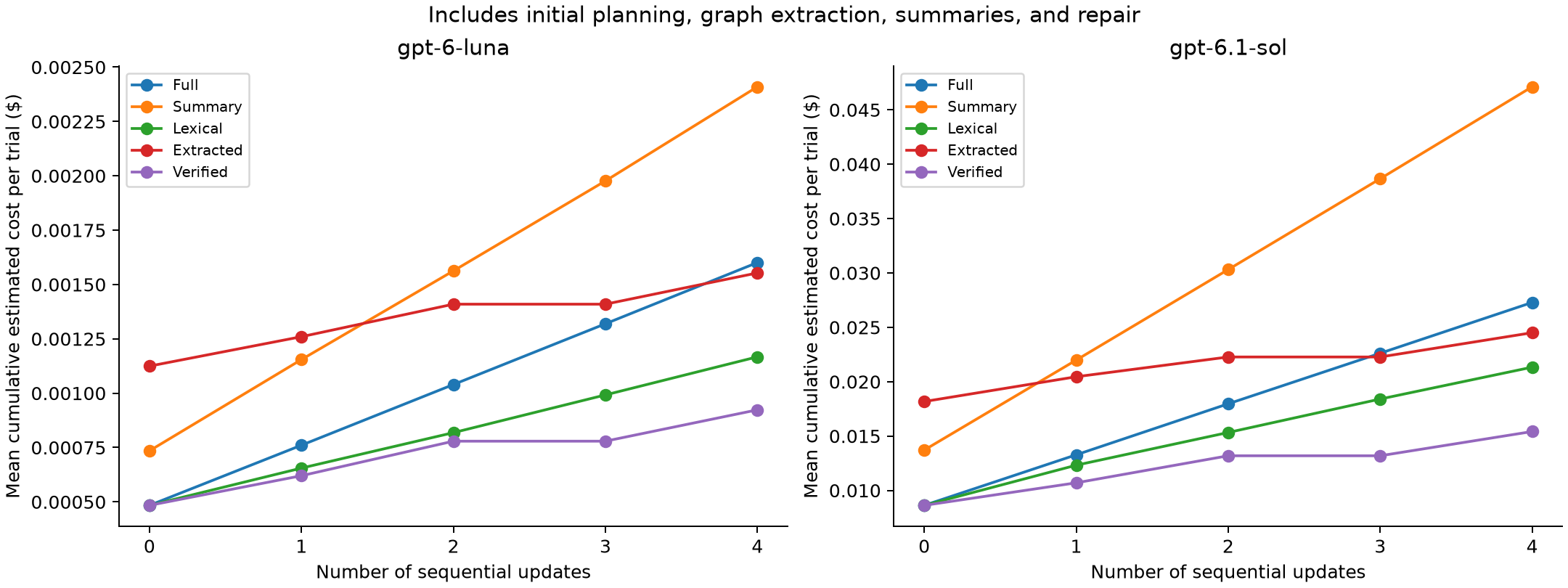}
\caption{Mean cumulative estimated cost, including initial agent calls and extraction or summary construction at update zero. Savings should be interpreted jointly with accuracy.}
\end{figure}

Initial plans are correct in 98.8\% of Luna trials and 100\% of Sol trials. Initial decision accuracy is 99.9\% and 100\%, respectively. Main extraction precision/recall is 99.9\%/94.9\% for Luna and 100\%/82.4\% for Sol. Relevant-record read coverage is 99.9\% and 92.6\%. Different read strategies produce different extraction evidence; these figures are not a controlled ranking of extraction capability independent of trace coverage.

\subsection{Dependency robustness}
On the first 30 main workflows per model, repetition zero, paired ablations remove field triples with a fixed SHA256 threshold at nominal 30\%, suppress transitive traversal, or add spurious dependencies. Table~\ref{tab:robust} shows large failures under missing edges and direct-only review.

The active spurious arm connects every output to the metadata field changed at update three, increasing unnecessary review. An initial implementation instead watched an unchanged contact-description field; that run is preserved as an inactive control, excluded from the active stress results. Its paid calls remain in project accounting. This correction affects the stress arm, not the frozen main comparison.
\begin{table}[htbp]
\centering\small
\begin{tabular}{llrr}
\toprule Model & Method & Accuracy (\%) & Reviewed\\\midrule
gpt-6-luna & Extracted graph & 88.3 & 5.87 \\
gpt-6-luna & Missing 30\% & 0.0 & 3.62 \\
gpt-6-luna & Active spurious & 90.0 & 9.53 \\
gpt-6-luna & Direct only & 9.2 & 2.15 \\
gpt-6.1-sol & Extracted graph & 70.8 & 5.42 \\
gpt-6.1-sol & Missing 30\% & 0.0 & 3.23 \\
gpt-6.1-sol & Active spurious & 71.7 & 9.08 \\
gpt-6.1-sol & Direct only & 9.2 & 2.10 \\
\bottomrule
\end{tabular}
\caption{Paired robustness subset: 30 workflow clusters per model, repetition zero.}
\label{tab:robust}
\end{table}

\subsection{Exploratory diagnosis}
Failure inspection revealed omitted alternative support and missing repair context. A procurement trace can establish that supplier A is initially eligible without reading suppliers B and C. Later, A may become ineligible; a graph that records only A can lack the fields needed to choose an alternative. A resource-allocation graph can omit the decision selecting which service's demand applies, leaving an allocation stale after order reversal.

Two exploratory arms use the same 30-workflow subset and initial plans. The first retains extracted-graph selection but supplies all current facts during repair. The second adds the complete record ID/field directory to the extraction input without revealing unread values or reference edges. Table~\ref{tab:explore} reports both. These comparisons were motivated by preliminary main errors, and are not preregistered primary comparisons. Separate model calls introduce sampling variation.
\begin{table}[htbp]
\centering\small
\begin{tabular}{llrr}
\toprule Model & Method & Accuracy (\%) & Total cost (\$)\\\midrule
gpt-6-luna & Selected outputs, full facts & 89.2 & 0.001753 \\
gpt-6.1-sol & Selected outputs, full facts & 100.0 & 0.027857 \\
gpt-6-luna & Directory-aware graph & 85.0 & 0.001575 \\
gpt-6.1-sol & Directory-aware graph & 100.0 & 0.026020 \\
\bottomrule
\end{tabular}
\caption{Exploratory context and extraction diagnostics on the paired subset.}
\label{tab:explore}
\end{table}

\subsection{Fresh-workflow confirmation}
After the exploratory diagnosis, the directory addition was frozen and evaluated on 60 additional workflows, generator indices 120--179, disjoint from the main set. There are 20 workflows per domain and two repeats per model, giving 240 initial trajectories and 1,920 method-update evaluations. These workflows share the original procedural generator, so this is fresh-instance confirmation rather than broad out-of-distribution validation.

\begin{table}[htbp]
\centering\small
\begin{tabular}{llrrrr}
\toprule Model & Method & Accuracy (\%) & 95\% CI & Reviewed & Total cost (\$)\\\midrule
gpt-6-luna & Full context & 99.6 & [99.0, 100.0] & 14.67 & 0.001600 \\
gpt-6-luna & Directory-aware graph & 84.8 & [79.4, 90.0] & 5.88 & 0.001577 \\
gpt-6.1-sol & Full context & 100.0 & [100.0, 100.0] & 14.67 & 0.027295 \\
gpt-6.1-sol & Directory-aware graph & 100.0 & [100.0, 100.0] & 6.08 & 0.026193 \\
\bottomrule
\end{tabular}
\caption{Fresh confirmation. The graph treatment receives the observable record directory during extraction.}
\label{tab:confirm}
\end{table}

For Sol, directory-aware graph repair matches observed full-context complete-plan correctness at 100\%, with estimated total cost 0.960 times full context (4.0\% lower). For Luna, accuracy is 84.8\% versus 99.6\%, a paired difference of $-14.8$ percentage points (95\% CI $[-20.2,-9.4]$), with cost 0.986 times full context. Thus the simple enhancement does not uniformly preserve correctness or deliver substantial quality-equivalent savings across models.

The all-success Sol intervals are degenerate. Even zero failing clusters among 60 independent workflows permits about a 4.9\% failure probability under a one-sided 95\% zero-event bound. We do not claim broad noninferiority or zero error.
\begin{figure}[htbp]
\centering\includegraphics[width=\linewidth]{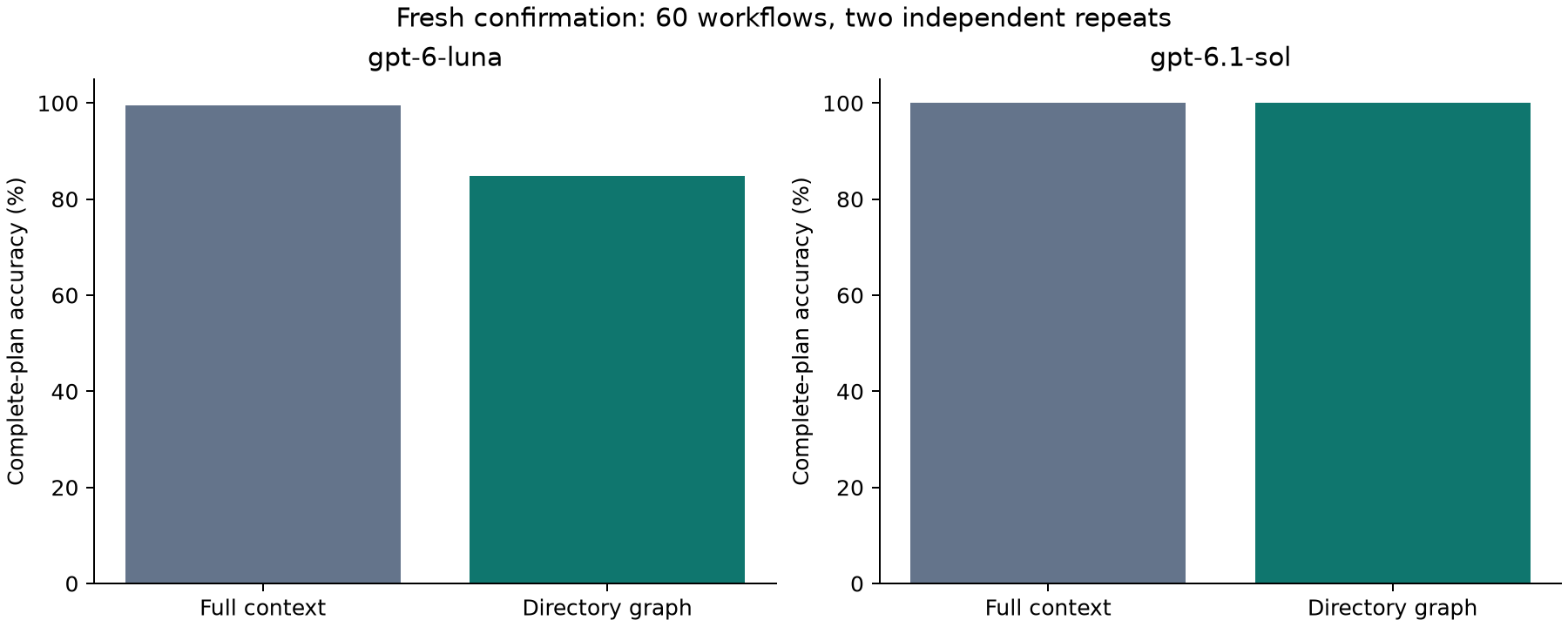}
\caption{Fresh-workflow confirmation of directory-aware extraction. The enhancement is model-dependent within this benchmark.}
\end{figure}

\section{Supplementary checks}
\subsection{LongMemEval evidence-only calibration}
We select 20 knowledge-update and 20 temporal-reasoning questions from the official cleaned LongMemEval oracle evidence-only file, ordered by SHA256(question ID), excluding abstention questions. Both models answer from the supplied dated evidence or a question-independent summary prompted to at most 4,000 characters. Reference answers and answer-session IDs are hidden during answering.

A Sol judge applies criteria adapted from the official evaluator, accepting equivalent updated answers and temporal off-by-one duration errors. There is no human judge calibration. Table~\ref{tab:external} reports these results as a separate answering/compression check. They are not full-history retrieval results, not leaderboard-comparable, and not evidence that graph repair generalizes to natural conversation.
\begin{table}[htbp]
\centering\small
\begin{tabular}{llrr}
\toprule Model & Type & Full evidence (\%) & Compressed evidence (\%)\\\midrule
gpt-6-luna & knowledge-update & 95.0 & 80.0 \\
gpt-6-luna & temporal-reasoning & 75.0 & 50.0 \\
gpt-6.1-sol & knowledge-update & 100.0 & 90.0 \\
gpt-6.1-sol & temporal-reasoning & 100.0 & 90.0 \\
\bottomrule
\end{tabular}
\caption{External evidence-only calibration, 20 questions per type and model. Scores are model-judged.}
\label{tab:external}
\end{table}

\subsection{Graph versus flat-list cost}
We measure 20 queries at each of 100, 1,000, and 10,000 nodes, with the same seven-decision affected chain. The prototype is compared with flat scans and rebuilt adjacency, including edge loading and index construction on every query. All return identical candidate sets. At 10,000 nodes, median query times are 23.55, 19.96, and 19.54 ms, respectively. The prototype loads all 8,750 edges despite the small affected set. Thus this implementation does not demonstrate an inherent speed advantage over equivalent flat dependencies. Persistent reverse indexing is a possible engineering improvement, not a measured result here. The microbenchmark uses local in-memory SQLite under concurrent model activity, not production storage latency.

\section{Discussion and limitations}
The main failure mechanism is a distinction between sufficient evidence for an initial decision and sufficient support coverage for future repair. Reverse reachability cannot recover edges or alternative facts that extraction omitted. Correct initial plans can therefore coexist with incomplete repair graphs. A record directory is a simple way to expose possible alternatives without leaking future values, but its benefit is not uniform across the tested models.

Total cost changes the efficiency conclusion. The Sol enhancement matches observed correctness on fresh instances, but saves only about 4\% over four updates after paying for initial planning and extraction. Low-cost review calls alone would suggest a stronger saving than the full pipeline supports. Conversely, oracle annotations substantially reduce estimated cost in the diagnostic bound, but their acquisition cost is excluded.

The dataset has explicit rules, modest contexts, one exact answer per output, four updates, and few procedural families. Imperative and expression references agree on generated states, but share a generator and have no independent human audit. The initial agent performs a record-reading action rather than rich long-horizon execution. The lexical baseline is weak and fixed, summaries have prompted limits, and close temporal/dependency methods are not reproduced. Static edge labels include possible alternative supports, not active causal necessity. Graph maintenance, action execution, partial observability recovery through new tool reads, and multiple acceptable repairs are not evaluated.

Model aliases and gateway prices are retained, but public dated weight snapshots are not established. Summed call durations under concurrency are not user-facing end-to-end latency. Workflow bootstrap intervals measure variation within this generator, not across organizations or task families. Multiple exploratory comparisons are disclosed without a broad adjusted significance claim. Real workflow traces, human annotation, stronger retrieval, close-baseline reproduction, and additional policy families are needed before broad agent-memory claims.

\section{Reproducibility and AI assistance}
The experiment retains exact prompts, public responses, tool observations, response IDs, served model identifiers, token usage, rates, elapsed times, source hashes, dataset hashes, and internally frozen protocols. Successful checkpoints are resumable without rebilling. No external preregistration is claimed. Across prior pilots, development, main, robustness, confirmation, and judge calls, the project completed 16,558 requests with no pending or uncertain calls. Rate-based usage is estimated at \$34.93; conservative admission accounting with a 25\% margin is \$43.66. Fifty targeted transport, budget, dependency, and workflow tests pass. No public code-repository availability is claimed in this version.

AI tools assisted software implementation, experiment orchestration, analysis, and manuscript drafting. The human author is responsible for the final manuscript and submission. AI-generated annotations are not described as independently human-verified. Generated workflow data contain fictional entities; this study does not involve real procurement, scheduling, or allocation decisions about people.

\section{Conclusion}
Trace-derived context graphs reduce review work but can fail when unvisited alternatives or selection-dependent fields are omitted. Directory-aware extraction recovers observed full-context correctness for one tested model on fresh procedural workflows, with a modest total-cost saving, while the other model remains less reliable. Selective repair should therefore be assessed using executable plan correctness, dependency coverage, and full memory-processing cost together. These results support a narrow empirical account of when selective repair fails and succeeds, rather than an unconditional graph-memory advantage.

\appendix
\section{Observable prompts and reference separation}
The initial request provides \texttt{policies} and a \texttt{record\_directory}, and requires a \texttt{read\_records} tool call. The returned records are followed by a request for every decision, value, and brief public explanation; insufficient observations allow null values. Extraction requests direct field-level dependencies and possible alternative supports. The main extraction input has policies, the initial plan, and actual read results. The directory-aware condition adds record IDs and field names, without unread values.

Repair requests say: recompute all requested decisions under the latest facts; return exactly the requested IDs; old values may be stale; compute requested dependent decisions together; return null if required information is absent. Inputs contain requested IDs, corresponding policies, latest retrieved context, prior decisions, and the current patch. Reference expression trees, reference outputs, and reference dependency labels are evaluator-only. The verified-graph diagnostic is the explicit exception for edge-based selection and retrieval, not for model answer generation.

\section{Frozen artifact identifiers}
Main generator SHA256: \path{11fb0caa00ea591a304b846241488bc72d808da685588e095d29bdfc6ac10753}. Main runner SHA256: \path{6699d13f88e7a02417ab8e8ae4166a49c52ac20139f2e41d6e3105164b4b225c}. Fresh confirmation dataset SHA256: \path{29657cc1f394e58db186dee679d8d6531ed29709414c8df44787f1c0bcb540fe}. Full trace archives are retained by the author; publishing a portable reproduction package is future work because gateway access and exact dated model snapshots are not public prerequisites established here.
\end{document}